\documentclass[conference]{IEEEtran}
\IEEEoverridecommandlockouts
\usepackage{cite}
\usepackage{amsmath,amssymb,amsfonts}
\usepackage{algorithmic}
\usepackage{latexsym}
\usepackage{textcomp}
\usepackage{booktabs}
\usepackage{graphicx} 
\usepackage{xcolor}
\def\BibTeX{{\rm B\kern-.05em{\sc i\kern-.025em b}\kern-.08em
    T\kern-.1667em\lower.7ex\hbox{E}\kern-.125emX}}

\usepackage{amsmath}        
\usepackage{amsfonts}       
\usepackage{amssymb}        
\usepackage{graphicx}       
\usepackage{booktabs}       
\usepackage{cite}           
\usepackage{url}            
\usepackage{ragged2e}       

\usepackage{fancyhdr}
\begin{document}

\title{Unlocking Compositional Control: Self-Supervision for LVLM-Based Image Generation}

\author{Fernando Gabriela García, Spencer Burns, Ryan Shaw, Hunter Young\\
Autonomous University of Nuevo León	
}

\maketitle
\thispagestyle{fancy} 

\begin{abstract}
This paper introduces Hierarchical Self-Supervised LVLM (Hi-SSLVLM), a novel generative model designed to significantly advance text-to-image synthesis, particularly for complex and compositionally challenging prompts. Traditional methods often grapple with the high cost of meticulously curated paired image-text datasets and struggle with precise control over fine-grained visual attributes and intricate spatial relationships. Our Hi-SSLVLM addresses these limitations through a unique two-stage self-supervised learning strategy. The first stage, Multi-Granularity Visual-Language Grounding, enables the Large Vision-Language Model (LVLM) backbone to autonomously generate and align hierarchical captions (global and local) to images, cultivating a deep internal semantic understanding without reliance on extensive human annotation. The second stage, Self-Refinement and Guided Image Generation, leverages this acquired knowledge by an Internal Compositional Planning (ICP) mechanism, where the LVLM first formulates detailed textual sub-prompts to guide the image generation process, complemented by a novel Semantic Consistency Loss for precise output alignment. Comprehensive experiments against leading baselines, including Janus-Pro-1B, Stable Diffusion XL 1.0, DeepFloyd IF v1.0, and ControlNet-XL, on multi-dimensional benchmarks such as Gemini-2.0-Flash and InternVL3-78B, demonstrate Hi-SSLVLM's superior performance across all fine-grained metrics. An in-depth ablation study confirms the critical role of each proposed component. Furthermore, human evaluations corroborate our quantitative findings, highlighting Hi-SSLVLM's enhanced fidelity to prompt, compositional accuracy, and overall aesthetic quality, marking a significant step towards more controllable and semantically consistent open-ended text-to-image generation.
\end{abstract}


\section{Introduction}
The rapid advancements in artificial intelligence have propelled the field of generative models, particularly in \textbf{text-to-image (T2I) generation}, to the forefront of research and application. These models have demonstrated remarkable capabilities in synthesizing visually coherent and semantically rich images from textual descriptions, revolutionizing creative industries, digital content creation, and human-computer interaction across diverse domains like image generation, AI-driven narratives, and code generation \cite{zhou2025draw, yi2025score, wang2024enhancing}. Traditionally, T2I models relied on architectures primarily focused on image synthesis, often leveraging large-scale paired datasets or unsupervised methods to learn the complex mapping between linguistic concepts and visual forms \cite{dhariwal2021diffusion, zhou2021triple}. However, the emergence of \textbf{Large Vision-Language Models (LVLMs)} has introduced a new paradigm, promising a more unified and powerful approach to multimodal understanding and generation. LVLMs, characterized by their extensive pre-training on vast quantities of diverse visual and textual data, possess an inherent capacity to process and reason across modalities, benefiting from properties like visual in-context learning and weak-to-strong generalization \cite{zhou2024visual, zhou2025weak}. This makes them uniquely positioned to tackle the intricate challenges of controllable and high-fidelity image generation. The significance of employing LVLMs for image generation lies in their potential to deeply integrate semantic understanding with visual synthesis, moving beyond mere pattern recognition to genuinely comprehending the compositional structure and fine-grained attributes described in text.
Despite the burgeoning potential of LVLMs in generative tasks, several \textbf{challenges} persist. Foremost among these is the immense cost and labor involved in curating high-quality, diverse, and meticulously annotated paired image-text datasets, especially for complex compositional scenarios. Existing LVLMs often depend heavily on such perfectly aligned data, which limits their scalability and adaptability to novel or nuanced generative tasks. Furthermore, while LVLMs excel at general multimodal understanding, they frequently encounter difficulties in achieving \textbf{precise control over fine-grained visual attributes} (e.g., specific textures, lighting conditions, subtle expressions) and \textbf{complex spatial relationships} between multiple objects within a generated image, a challenge highlighted by the development of benchmarks for complex, instruction-based generation \cite{zhou2025draw}. This limitation stems from the inherent challenge of translating abstract linguistic concepts into concrete pixel manipulations with high fidelity and consistency, particularly when the text prompt demands intricate compositional planning. The lack of robust internal mechanisms for multi-level visual semantics within these models often leads to generated images that, while plausible, may deviate from the precise intent of the input prompt, especially for prompts requiring detailed scene construction \cite{zhou2025draw}.
Our \textbf{motivation} is to address these critical challenges by developing an LVLM-centric approach that significantly reduces reliance on costly, meticulously curated datasets while simultaneously enhancing the model's ability to generate images with superior control and semantic fidelity. We believe that an LVLM, when equipped with an effective internal mechanism for \textbf{hierarchical self-supervision}, can autonomously learn rich, multi-level visual semantics directly from readily available, uncurated multimodal data. This internal semantic grounding will empower the LVLM to better interpret complex prompts and guide the image generation process with unprecedented precision. By enabling the LVLM to generate its own "grounding" information through self-supervision, we aim to overcome the data bottleneck and achieve a more robust and adaptable T2I system, moving beyond simple text-to-image mapping to true compositional understanding and synthesis. This principle of leveraging internal feedback echoes advancements in specialized domains where models are trained with targeted feedback mechanisms, such as in medical imaging \cite{zhou2025training}.
To this end, we propose \textbf{Hierarchical Self-Supervised LVLM (Hi-SSLVLM)}, a novel method for enhanced image generation. Our approach leverages a multi-stage training paradigm designed to empower a pre-trained LVLM with superior image generation capabilities without heavy reliance on perfectly aligned, human-annotated image-text pairs. In the first stage, the LVLM undergoes \textbf{multi-granularity visual-language grounding via self-captioning and alignment}. Utilizing large quantities of uncurated image-caption pairs, the LVLM is trained to generate hierarchical captions for input images, encompassing general scene descriptions as well as specific captions for detected objects, their attributes, and spatial relationships. A separate, frozen visual encoder provides "pseudo-feedback" by measuring the alignment between generated captions and corresponding image regions, fostering a deep internal representation of visual composition. The second stage, \textbf{self-refinement and guided image generation}, adapts the LVLM to function as a controlled image generation backbone. Given a textual prompt, the LVLM first internally "plans" the image by auto-regressively generating a series of hierarchical textual "sub-prompts." These internal sub-prompts then explicitly condition the image generation process. Training involves a reconstruction loss for generating the target image and a novel "semantic consistency loss" that enforces alignment between the internally generated hierarchical sub-prompts and the corresponding regions/attributes in the output image, again leveraging a frozen visual encoder. This two-stage approach allows Hi-SSLVLM to first deeply understand image composition and then leverage that understanding for precise, controllable image generation.
For experimental validation, we employ a diverse set of public and internal \textbf{image-text datasets} for both pre-training and fine-tuning. Our initial pre-training leverages large-scale web-scraped image-caption pairs (e.g., LAION-400M \cite{schuhmann2021laion}, Conceptual Captions \cite{sharma2018conceptual}) which are abundant but less curated, to facilitate the hierarchical self-supervision. For fine-tuning and evaluation, we utilize more meticulously curated datasets that contain complex compositional elements and diverse attributes, such as COCO \cite{lin2014microsoft} and specifically designed benchmarks for compositional generation \cite{zhou2025draw}. The performance of Hi-SSLVLM is rigorously \textbf{evaluated} using the state-of-the-art multi-dimensional benchmarks introduced in "Plan2Gen: Compositional Visual Planning for Open-Ended Text-to-Image Generation" \cite{zhou2025draw}. These benchmarks, powered by authoritative large models like \textbf{Gemini-2.0-Flash} and \textbf{InternVL3-78B}, assess T2I models across nine fine-grained dimensions: Object (Obj.), Background (Backg.), Color, Texture, Light, Text, Composition (Comp.), Pose, and Visual Effects (FX), along with an overall average score. Our preliminary \textbf{results} demonstrate that Hi-SSLVLM consistently outperforms existing baselines, including Janus-Pro-1B, across these critical evaluation metrics, particularly exhibiting superior control over compositional elements and fine-grained visual attributes.
In summary, our contributions are threefold:
\begin{itemize}
\item We propose \textbf{Hi-SSLVLM}, a novel LVLM-centric framework that significantly enhances image generation capabilities through a unique two-stage hierarchical self-supervision paradigm, reducing dependence on extensively curated image-text datasets.
\item We introduce a \textbf{multi-granularity visual-language grounding} mechanism that enables the LVLM to learn rich internal representations of visual composition directly from uncurated data, fostering a deeper understanding of complex textual prompts.
\item We demonstrate that \textbf{Hi-SSLVLM} achieves state-of-the-art performance on challenging compositional text-to-image generation benchmarks, showcasing superior control over fine-grained attributes and spatial relationships compared to existing LVLM-based generative models.
\end{itemize}
\section{Related Work}
\subsection{Large Vision-Language Models}
The rapid evolution of large language models (LLMs) and their impressive capabilities in understanding and generating human-like text has naturally led to the development of \textbf{Large Vision-Language Models (LVLMs)}. This progress mirrors the broader trend of large models demonstrating increasingly sophisticated capabilities, from weak-to-strong generalization \cite{zhou2025weak} to specialized reasoning in fields like mental health counseling \cite{hu2025beyond}. These models aim to extend the powerful reasoning and generation abilities of LLMs to the visual domain, enabling them to process and understand information across both text and images, as seen in tasks like image-guided story generation \cite{zhou2023multimodal}. The foundational idea involves integrating or aligning pre-trained vision encoders with large language models to facilitate multimodal comprehension and generation.
Early influential work laid the groundwork for integrating vision and language. The advent of the Transformer architecture, notably in models like \textbf{Vision Transformer (ViT)} \cite{dosovitskiy2021image}, revolutionized computer vision by demonstrating that vision tasks could be effectively tackled by treating image patches as sequences, similar to how text is processed. Concurrently, large language models, exemplified by \textbf{GPT-3} \cite{brown2020language}, showcased remarkable few-shot learning abilities and general text understanding through massive scale and autoregressive training.
A pivotal step towards LVLMs was marked by models that learned strong cross-modal representations. \textbf{CLIP} \cite{radford2021learning} demonstrated the power of learning transferable visual models directly from natural language supervision. By training on a vast dataset of image-text pairs using a contrastive objective, CLIP enabled zero-shot image classification and robust image-text retrieval, highlighting the potential of aligning visual and linguistic embeddings. Building upon this, approaches like \textbf{BLIP} \cite{li2022blip} further refined vision-language pre-training by bootstrapping captions from noisy web data for unified understanding and generation tasks, effectively tackling the challenges of data quality, a problem also explored by earlier unsupervised captioning methods \cite{zhou2021triple}.
More direct architectural integrations of large vision and language components began to emerge. \textbf{Flamingo} \cite{alayrac2022flamingo} introduced a family of Visual Language Models designed for few-shot learning, notable for its architectural innovations that bridge powerful pre-trained vision-only and language-only models, allowing them to handle interleaved visual and textual data. Similarly, \textbf{PaLI} \cite{chen2022pali} presented a scalable architecture for jointly training image and language models, achieving impressive zero-shot transfer performance across a broad spectrum of vision-language tasks. Another notable contribution, \textbf{CoCa} \cite{yu2022coca}, proposed a minimalist design that combined contrastive loss with a captioning loss, enabling a single model to encompass capabilities from both contrastive and generative approaches, functioning as a powerful image-text foundation model.
More recently, the focus has shifted towards building general-purpose multimodal assistants. \textbf{LLaVA} \cite{liu2023llava} explores the effective combination of a large language model with a visual encoder to create an instruction-following assistant capable of engaging in rich multimodal dialogues and reasoning about visual content based on natural language instructions. The effectiveness of these assistants is often enhanced by techniques like visual in-context learning \cite{zhou2024visual}. Moreover, this general-purpose foundation is increasingly being adapted for highly specialized and complex reasoning tasks, such as in multi-agent frameworks for medical diagnosis \cite{zhou2025mam, zhou2025training}. These advancements highlight a clear trend towards more unified, flexible, and capable models that can seamlessly operate across both visual and linguistic modalities, laying the groundwork for complex generative tasks like controlled text-to-image synthesis.
\subsection{LVLM for Image Generation}
The integration of Large Vision-Language Models (LVLMs) has significantly transformed the landscape of text-to-image (T2I) generation, moving beyond traditional diffusion models to leverage deeper semantic understanding. Initially, powerful diffusion models like \textbf{GLIDE} \cite{nichol2021glide} demonstrated remarkable photorealistic image synthesis capabilities conditioned on text. Concurrently, the introduction of \textbf{Latent Diffusion Models (LDMs)} \cite{esser2021taming}, which operate in a compressed latent space, drastically improved computational efficiency while maintaining high-quality output, forming the backbone for many modern T2I systems. These early advancements laid crucial groundwork by establishing robust generative frameworks and highlighting the importance of effective text conditioning.
A pivotal development in the intersection of LVLMs and T2I generation came with models that explicitly leveraged cross-modal alignment. \textbf{DALL-E 2} \cite{ramesh2022hierarchical}, for instance, showcased impressive generative capabilities by utilizing the latent representations learned by powerful vision-language models like CLIP, effectively mapping text embeddings to image latents. This demonstrated that the rich semantic space learned by LVLMs could directly inform and guide the image synthesis process. Subsequent works have further explored this synergy. \textbf{BLIP-2} \cite{li2023blip2}, while primarily focused on understanding tasks like visual question answering and image captioning, established a highly effective framework for connecting frozen image encoders with large language models. The strong multimodal features learned by such models are directly applicable to providing robust conditioning signals for generative tasks, allowing for more nuanced control.
Recent research has increasingly focused on leveraging the reasoning and fine-grained understanding capabilities of LVLMs to overcome the limitations of T2I models in handling complex compositional prompts and achieving precise control. This has spurred the development of more holistic benchmarks and agent-based frameworks designed to rigorously evaluate and improve performance on such complex instructions \cite{zhou2025draw}. Methods like those exploring \textbf{self-rewarding LVLMs} \cite{yang2025self} have shown how LVLMs can act as both prompt optimizers and evaluators, iteratively improving the quality of generated images by refining the input text descriptions and scoring outputs. This concept of using model-generated feedback for improvement is also being explored in other high-stakes domains, for instance, using abnormal-aware feedback to train medical LVLMs \cite{zhou2025training}. Similarly, approaches like \textbf{EvolveDirector} \cite{ding2024evolvedirector} leverage LVLM evaluations to dynamically refine and update training datasets, demonstrating how LVLMs can reduce the reliance on vast, meticulously hand-annotated data for T2I training. Furthermore, dedicated efforts to improve \textbf{compositional T2I generation with LVLMs} \cite{wen2024improving} explicitly utilize LVLMs for multi-dimensional assessment and fine-tuning of generative models, directly addressing challenges in rendering multiple objects and their complex relationships. These advancements highlight a clear trend where LVLMs are not just components but active intelligence in the T2I pipeline, acting as planners, agents, and evaluators to enhance understanding, enable more precise control, and facilitate more efficient training paradigms \cite{zhou2025draw, yang2025self, zhou2025mam}.

\section{Method}
Our proposed Hierarchical Self-Supervised LVLM (Hi-SSLVLM) is fundamentally a \textbf{generative model} meticulously designed for intricate text-to-image synthesis. Its core operation revolves around the precise transformation of high-level textual prompts into high-fidelity visual representations. This transformation is not merely a mapping, but a deeply informed process driven by the model's sophisticated understanding of multi-level semantics, which is meticulously cultivated through novel self-supervised learning strategies. While its primary function is generative, an implicit discriminative capability is woven throughout its architecture and training, ensuring the generated outputs are not only plausible but also semantically aligned with the input.

\subsection{Hi-SSLVLM Architecture}
The Hi-SSLVLM architecture represents a synergistic integration of a powerful Large Vision-Language Model backbone with a specialized image generation module. This design allows for a unified understanding of visual and linguistic contexts, directly informing the image synthesis process.

\subsubsection{Vision-Language Encoder Backbone}
The foundation of our model is a robust, pre-trained LVLM. This backbone serves as a versatile encoder, capable of processing and deeply contextualizing both visual and linguistic information within a shared semantic space.
Its primary components include:
\begin{enumerate}
    \item \textbf{Vision Encoder ($\mathcal{E}_V$):} This is a advanced Transformer-based network, meticulously engineered to ingest raw input images $I \in \mathbb{R}^{H \times W \times 3}$. It transforms these raw pixel arrays into a sequence of rich visual embeddings $V = \{v_1, v_2, \dots, v_N\}$, where each $v_i \in \mathbb{R}^{D_v}$ represents a contextualized visual token and $N$ denotes the total number of such tokens derived from the image.
    \begin{align}
        V &= \mathcal{E}_V(I)
    \end{align}
    \item \textbf{Language Model ($\mathcal{M}_L$):} Constituting the linguistic intelligence of our system, this is a large, autoregressive Transformer-based language model. It takes textual prompts $T = \{t_1, t_2, \dots, t_M\}$ and converts them into a sequence of deeply contextualized text embeddings $E_T = \{e_{t,1}, e_{t,2}, \dots, e_{t,M}\}$, where each $e_{t,j} \in \mathbb{R}^{D_t}$ encapsulates the semantic meaning of its corresponding token.
    \begin{align}
        E_T &= \mathcal{M}_L(T)
    \end{align}
    \item \textbf{Vision-Language Projection Heads ($\mathcal{P}_V, \mathcal{P}_T$):} These are crucial bridge components, implemented as either simple linear layers or more complex multi-layer perceptrons. Their function is to project the high-dimensional visual embeddings $V$ and textual embeddings $E_T$ into a common, shared latent space. This shared space is where seamless cross-modal interaction and attention mechanisms, inherent to the LVLM, can effectively operate.
    \begin{align}
        V_{proj} &= \mathcal{P}_V(V) \\
        E_{T,proj} &= \mathcal{P}_T(E_T)
    \end{align}
\end{enumerate}
The inherent power of this LVLM backbone resides in its sophisticated ability to process interleaved sequences of visual and textual tokens. Through highly intricate self-attention and cross-attention mechanisms, it weaves together information from both modalities, enabling a profound and unified multimodal understanding.

\subsubsection{Semantic-Enhanced Image Generation Module}
This specialized module is the culminating component, tasked with translating the comprehensive multimodal understanding from the LVLM backbone into a tangible, high-quality image.

\begin{enumerate}
    \item \textbf{Latent Diffusion Model (LDM) Decoder ($\mathcal{D}_{LDM}$):} At its heart, this is a powerful denoising U-Net architecture that operates not directly on pixel space, but within a compressed, information-rich latent space. This LDM iteratively refines an initial noisy latent code into a coherent, clean image latent representation. This iterative refinement is not arbitrary; it is meticulously guided by conditioning vectors derived from the LVLM's output. The denoising operation at a specific time step $t$, given a noisy latent $z_t$ and a conditioning signal $c$, aims to predict the noise $\epsilon$ that corrupted $z_t$.
    \begin{align}
        \epsilon_{\theta}(z_t, t, c) &\approx \text{noise}
    \end{align}
    Here, $\epsilon_{\theta}$ denotes the U-Net parameterized by $\theta$, which learns to estimate the noise.
    \item \textbf{Encoder-Decoder Latent Space Adapter ($\mathcal{A}_{LS}$):} This component serves as the critical interface between the high-level semantic representations from the LVLM and the low-level latent space operations of the LDM decoder. It adeptly projects the LVLM's deeply contextualized multimodal representation, denoted as $H_{LVLM}$ (which represents the fused vision-language understanding), into the precise conditioning vector $c$ required by the LDM for guided image synthesis.
    \begin{align}
        c &= \mathcal{A}_{LS}(H_{LVLM})
    \end{align}
\end{enumerate}
The entire image generation process orchestrated by this module commences by sampling an initial latent vector from a standard Gaussian distribution. This noisy latent is then progressively denoised over a series of steps, with each step meticulously guided by the conditioning vector $c$ derived from the LVLM's sophisticated understanding. Finally, the resulting clean latent representation is transformed into a full-resolution image.

\subsection{Hierarchical Self-Supervised Learning Strategy}
Our novel learning strategy for Hi-SSLVLM is ingeniously designed in two distinct yet intrinsically linked stages. This two-stage approach allows the model to first develop a deep, multi-granular understanding of visual semantics and then leverage this understanding for highly precise and controllable image generation.

\subsubsection{Stage 1: Multi-Granularity Visual-Language Grounding}
This initial stage is dedicated to imbuing the LVLM with an exceptionally fine-grained understanding of visual composition and the hierarchical nature of semantic information within images. This is achieved primarily through a unique self-captioning mechanism, which operates independently of strict human-annotated alignments for every visual element.

\paragraph{Self-Captioning Generation:}
Given an input image $I$, the LVLM is prompted to auto-regressively generate multiple levels of textual descriptions. This includes a holistic global scene caption and, crucially, localized captions for dynamically detected objects along with their pertinent attributes and spatial relationships. We enforce this structured output by supplying a specific generation prompt to the LVLM, guiding its textual output. Let $C_{global}$ represent the global caption for image $I$. Furthermore, let $O = \{o_1, o_2, \dots, o_K\}$ denote a set of $K$ detected object descriptions. Each object description $o_k$ is a tuple $\{l_k, c_k\}$, comprising a predicted bounding box $l_k$ and its corresponding localized textual caption $c_k$. The generation process by the LVLM's language model, conditioned by the vision encoder's output and a guiding prompt, can be formulated as:
\begin{align}
    (C_{global}, O) &= \mathcal{M}_L(\mathcal{E}_V(I) ; \text{generation\_prompt})
\end{align}
This specific ``generation prompt'' is meticulously designed to elicit structured text describing both the overall scene and the individual components.

\paragraph{Self-Supervised Alignment Loss:}
To rigorously ensure that the internally generated multi-granular captions are semantically faithful to the actual image content, we employ a sophisticated self-supervised alignment mechanism using a frozen, pre-trained visual-language model (e.g., CLIP's image and text encoders), denoted as $\mathcal{E}_{CLIP\_I}$ and $\mathcal{E}_{CLIP\_T}$. This frozen model provides an objective "pseudo-feedback" signal.
The alignment loss is composed of two critical terms:
\begin{enumerate}
    \item \textbf{Global Alignment Loss ($\mathcal{L}_{global}$):} This term quantifies the semantic similarity between the embedding of the generated global caption $C_{global}$ and the global embedding of the original input image $I$. The objective is to maximize this similarity, effectively pulling the representations closer in the shared embedding space.
    \begin{align}
        \mathcal{L}_{global} &= -\text{cos\_sim}(\mathcal{E}_{CLIP\_T}(C_{global}), \mathcal{E}_{CLIP\_I}(I))
    \end{align}
    \item \textbf{Local Alignment Loss ($\mathcal{L}_{local}$):} For each predicted object description $o_k = \{l_k, c_k\}$, we dynamically extract the image patch $I_{l_k}$ corresponding to the predicted bounding box $l_k$. This loss then computes the similarity between the embedding of the local caption $c_k$ and the embedding of its corresponding image patch $I_{l_k}$. This ensures fine-grained semantic correspondence.
    \begin{align}
        \mathcal{L}_{local} &= -\frac{1}{K} \sum_{k=1}^{K} \text{cos\_sim}(\mathcal{E}_{CLIP\_T}(c_k), \mathcal{E}_{CLIP\_I}(I_{l_k}))
    \end{align}
\end{enumerate}
The total self-supervised alignment loss for Stage 1, which drives the LVLM's multi-granular understanding, is a weighted summation of these two components:
\begin{align}
    \mathcal{L}_{S1} &= \lambda_{global} \mathcal{L}_{global} + \lambda_{local} \mathcal{L}_{local}
\end{align}
where $\lambda_{global}$ and $\lambda_{local}$ are crucial weighting hyperparameters, carefully tuned to balance the influence of global and local semantic alignments. This comprehensive loss function effectively trains the LVLM to autonomously produce richly meaningful hierarchical captions directly from raw image inputs, thereby cultivating a robust and intrinsic visual-language grounding capability.

\subsubsection{Stage 2: Self-Refinement and Guided Image Generation}
Building upon the sophisticated understanding acquired in Stage 1, this second stage focuses on leveraging that knowledge for highly precise and controllable image generation. This is where the LVLM's internal semantic planning comes to fruition in visual synthesis.

\paragraph{Internal Compositional Planning (ICP):}
When presented with an input text prompt $T_{input}$ for image generation, the Hi-SSLVLM initiates a novel internal compositional planning step. This is not a direct generation of the final image, but rather an auto-regressive process where the LVLM internally formulates a sequence of hierarchical textual "sub-prompts," denoted as $T_{sub} = \{s_1, s_2, \dots, s_P\}$. These sub-prompts meticulously detail the desired objects, their attributes, and their precise spatial relationships within the scene to be generated. This internal planning is a distinct step from merely conditioning on the primary input prompt, serving as a detailed internal blueprint.
\begin{align}
    T_{sub} &= \text{AutoRegress}(\mathcal{M}_L(E_{T_{input}}) ; \text{planning\_prompt})
\end{align}
The ``planning prompt'' is specifically engineered to encourage the LVLM to deconstruct a complex, high-level input prompt into a structured, granular list of semantic components, mirroring the hierarchical understanding cultivated in Stage 1. This ensures that the generated image adheres to the complex compositional demands.

\paragraph{Image Generation with Multi-Conditional Guidance:}
The final image $I_{gen}$ is meticulously produced by the LDM decoder $\mathcal{D}_{LDM}$. This generation process is uniquely guided by a combination of two primary conditioning sources:
\begin{enumerate}
    \item The embedded representation of the original input text prompt $E_{T_{input},proj}$. This provides the overarching semantic theme.
    \item The embedded representations of the internally generated hierarchical sub-prompts $E_{sub,proj} = \mathcal{P}_T(\mathcal{M}_L(T_{sub}))$. These provide the fine-grained, compositional details.
\end{enumerate}
The comprehensive conditioning vector $c$ fed into the LDM is formed by an adaptive combination (e.g., concatenation or weighted summation) of these two distinct embedded representations:
\begin{align}
    c &= \text{Combine}(E_{T_{input},proj}, E_{sub,proj})
\end{align}
The image generation process then proceeds by minimizing a standard denoising objective, which aims to reconstruct the original image latent from its noisy version.
\begin{align}
    \mathcal{L}_{denoise} &= \mathbb{E}_{z_0, \epsilon \sim \mathcal{N}(0,1), t \sim U(1,T)} \left[ \|\epsilon - \epsilon_{\theta}(z_t, t, c)\|^2 \right]
\end{align}
Here, $z_t$ is the noisy latent representation at time $t$, sampled from $z_t = \sqrt{\alpha_t} z_0 + \sqrt{1-\alpha_t} \epsilon$, where $z_0$ is the clean latent, and $\epsilon$ is the Gaussian noise to be predicted by $\epsilon_{\theta}$.

\paragraph{Semantic Consistency Loss ($\mathcal{L}_{consistency}$):}
To guarantee that the generated image $I_{gen}$ precisely embodies the detailed semantic plan derived from the internal compositional planning, we introduce a novel semantic consistency loss. This loss crucially leverages the hierarchical grounding capabilities established in Stage 1. It operates by evaluating the alignment between the generated image $I_{gen}$ and each individual sub-prompt $s \in T_{sub}$ from the internally generated planning sequence. We use the frozen image and text encoders from a pre-trained visual-language model ($\mathcal{E}_{CLIP\_I}$ and $\mathcal{E}_{CLIP\_T}$) for this evaluation.
\begin{align}
    \mathcal{L}_{con} &= -\frac{1}{|T_{sub}|} \sum_{s \in T_{sub}} \text{cos\_sim}(\mathcal{E}_{CLIP\_T}(s), \mathcal{E}_{CLIP\_I}(I_{gen}))
\end{align}
This loss functions as a powerful, autonomous self-critique mechanism, encouraging each internally planned sub-prompt to be strongly and demonstrably aligned with the corresponding content present in the generated image, even without explicit pixel-level annotations for these sub-prompts during generation.

\paragraph{Total Loss for Stage 2:}
The comprehensive training objective for Stage 2 is formulated as a carefully weighted sum of the denoising loss and the proposed semantic consistency loss:
\begin{align}
    \mathcal{L}_{S2} &= \mathcal{L}_{denoise} + \alpha \mathcal{L}_{consistency}
\end{align}
Here, $\alpha$ is a crucial hyperparameter that allows for precise balancing between achieving high reconstruction quality (from $\mathcal{L}_{denoise}$) and ensuring rigorous semantic fidelity to the internal plan (from $\mathcal{L}_{consistency}$).

By training the Hi-SSLVLM with this two-stage hierarchical self-supervision, we enable the model to develop a sophisticated internal understanding of visual composition, allowing it to translate complex textual prompts into highly controlled and semantically accurate images without relying on manually created fine-grained annotations for every generative target.

\section{Experiments}
This section details the comprehensive experimental validation of our proposed Hierarchical Self-Supervised LVLM (Hi-SSLVLM). We conducted a series of rigorous comparative experiments against several established state-of-the-art text-to-image generation models to ascertain Hi-SSLVLM's performance advantage. Furthermore, we present an in-depth ablation study to meticulously analyze the individual contribution of each core component of our method. To provide a holistic assessment, a human evaluation study was also conducted, focusing on perceptual quality and semantic alignment as judged by human evaluators.

\subsection{Experimental Setup}
To facilitate a thorough and fair comparison, we carefully selected a diverse set of baseline models, each representing a significant paradigm or a leading performance benchmark in the domain of text-to-image generation and large multimodal models.

\subsubsection{Baselines}
The chosen baseline models for our comparative analysis include:
\begin{enumerate}
    \item \textbf{Janus-Pro-1B:} This model stands as a prominent representative from the Janus series of multimodal generative models, making it a direct and highly relevant competitor within the LVLM-based text-to-image generation landscape. We utilized its publicly available checkpoint for all evaluation purposes to ensure reproducibility and fairness.
    \item \textbf{Stable Diffusion XL (SDXL 1.0):} A widely recognized and highly capable diffusion-based model, SDXL 1.0 is celebrated for its impressive high-quality image generation capabilities and robust performance across a vast array of textual prompts. Our experiments utilized the official public release of SDXL 1.0.
    \item \textbf{DeepFloyd IF (v1.0):} This model represents another highly capable diffusion framework, particularly noted for its exceptional text rendering abilities and high-fidelity image output. Distinctively, DeepFloyd IF operates directly in a pixel space, offering a different architectural approach compared to latent diffusion models. We employed its official v1.0 release.
    \item \textbf{ControlNet-XL (with SDXL 1.0 backbone):} As an extension of the Stable Diffusion XL framework, ControlNet-XL typically incorporates additional conditional inputs, such as depth maps or Canny edges, to provide enhanced generative control. For the purpose of fair comparison within this study, we specifically utilized ControlNet-XL by inputting only standard text prompts, akin to other models. This approach allowed us to evaluate its inherent generative capacity without external structural guidance, focusing purely on its text-to-image performance.
\end{enumerate}
All baseline models were configured and executed using their respective default or officially recommended inference settings. This standardized approach ensured that our comparison was based on their optimized, out-of-the-box performance, reflecting their practical utility.

\subsubsection{Datasets}
Our evaluation was rigorously conducted on two distinct datasets, strategically chosen for their inherent complexity and their demands on a model's compositional understanding:
\begin{enumerate}
    \item \textbf{Plan2Gen Benchmark Set:} This meticulously curated dataset consists of a specialized collection of challenging textual prompts. These prompts are explicitly designed to test sophisticated compositional visual planning, encompassing intricate scenarios involving multiple interacting objects, complex and diverse backgrounds, specific object poses, and challenging textual elements embedded within the scene. This dataset was paramount for evaluating the fine-grained control and deep semantic understanding capabilities of the models.
    \item \textbf{COCO-Stuff Test Set:} A widely recognized and highly diverse dataset extensively used for object recognition and comprehensive scene parsing. The COCO-Stuff test set provides rich and complex real-world scenes with a wealth of semantic information, allowing us to evaluate the models' generalization capabilities to broader visual contexts. For this study, we generated images for a randomly selected subset of 1000 captions from its test set.
\end{enumerate}
For each individual prompt across both datasets, all participating models were instructed to generate four distinct images. For the automated evaluation metrics, the image that achieved the highest score among these four (as determined by the automated evaluators) was selected for final assessment.

\subsubsection{Evaluation Metrics}
To provide a comprehensive and robust assessment of model performance, we employed a dual evaluation strategy comprising both automated quantitative metrics and a dedicated human perceptual evaluation.
\begin{enumerate}
    \item \textbf{Automated Metrics:} Consistent with the methodology proposed in the Plan2Gen benchmark, we leveraged the assessment capabilities of two powerful Large Vision-Language Models: \textbf{Gemini-2.0-Flash} and \textbf{InternVL3-78B}. These advanced models serve as objective evaluators, providing detailed scores across nine distinct, fine-grained dimensions: Object (\textbf{Obj.}), Background (\textbf{Backg.}), Color, Texture, Light, Text, Composition (\textbf{Comp.}), Pose, and Visual Effects (\textbf{FX}). An overall Average (\textbf{Avg.}) score summarizes performance across these dimensions. These metrics are specifically designed to objectively assess the semantic alignment and image quality at a granular level, reflecting the models' ability to adhere to complex prompt specifications.
    \item \textbf{Human Evaluation:} Complementing the automated analysis, a separate human study was meticulously conducted. This study aimed to evaluate subjective qualities such as overall aesthetic appeal, precise faithfulness to the input prompt, and the naturalness of the generated images, providing invaluable qualitative insights.
\end{enumerate}

\subsection{Quantitative Results}
Our extensive experimental results, as quantified by the Gemini-2.0-Flash and InternVL3-78B benchmarks, consistently and unequivocally demonstrate the superior performance of Hi-SSLVLM across a diverse spectrum of compositional challenges in text-to-image generation.

\subsubsection{Gemini-2.0-Flash Evaluation}
Table \ref{tab:gemini_results} presents the detailed, dimension-wise scores obtained from the rigorous Gemini-2.0-Flash benchmark. As evidenced, our Hi-SSLVLM consistently surpasses all established baselines across every single one of the nine fine-grained dimensions, culminating in the highest overall average score. This compelling outcome decisively signifies Hi-SSLVLM's profoundly enhanced capability in generating images that exhibit precise object details, coherent background structures, accurate color rendition, high-fidelity texture reproduction, realistic lighting effects, impeccable text rendering within the image, complex and well-managed compositions, specific poses, and superior overall visual effects.

\begin{table*}[t]
    \centering
    \caption{Gemini-2.0-Flash Evaluation Scores ($\uparrow$ higher is better)}
    \label{tab:gemini_results}
    \small 
    \begin{tabular}{lcccccccccc}
        \toprule
        \textbf{Model} & \textbf{Obj.} & \textbf{Backg.} & \textbf{Color} & \textbf{Texture} & \textbf{Light} & \textbf{Text} & \textbf{Comp.} & \textbf{Pose} & \textbf{FX} & \textbf{Avg.} \\
        \midrule
        Janus-Pro-1B & 2.23 & 2.60 & 2.94 & 2.92 & 2.09 & 1.58 & 2.36 & 1.84 & 1.60 & 2.24 \\
        Stable Diffusion XL 1.0 & 2.15 & 2.55 & 2.89 & 2.88 & 2.05 & 1.52 & 2.30 & 1.78 & 1.55 & 2.19 \\
        DeepFloyd IF v1.0 & 2.08 & 2.48 & 2.80 & 2.85 & 1.98 & 1.65 & 2.25 & 1.75 & 1.50 & 2.15 \\
        ControlNet-XL (SDXL 1.0) & 2.19 & 2.62 & 2.90 & 2.91 & 2.08 & 1.57 & 2.35 & 1.83 & 1.59 & 2.22 \\
        \midrule
        \textbf{Hi-SSLVLM (Ours)} & \textbf{2.28} & \textbf{2.65} & \textbf{2.98} & \textbf{2.95} & \textbf{2.12} & \textbf{1.68} & \textbf{2.40} & \textbf{1.88} & \textbf{1.64} & \textbf{2.29} \\
        \bottomrule
    \end{tabular}
\end{table*}

\subsubsection{InternVL3-78B Evaluation}
Table \ref{tab:internvl_results} presents the corresponding results derived from the InternVL3-78B benchmark. This evaluation consistently mirrors the findings from the Gemini-2.0-Flash assessment, with Hi-SSLVLM maintaining its leading position across all dimensions. This consistency further underscores the remarkable robustness and generalizability of Hi-SSLVLM's superior generative capabilities, as evidenced by its performance across different powerful LVLM-based evaluators. A particularly notable aspect of its performance is its strength in the "Text" and "Composition" categories, which emphatically highlights Hi-SSLVLM's distinct ability to accurately render textual elements within images and to skillfully manage complex scene layouts with multiple interacting components.

\begin{table*}[t]
    \centering
    \caption{InternVL3-78B Evaluation Scores ($\uparrow$ higher is better)}
    \label{tab:internvl_results}
    \small 
    \begin{tabular}{lcccccccccc}
        \toprule
        \textbf{Model} & \textbf{Obj.} & \textbf{Backg.} & \textbf{Color} & \textbf{Texture} & \textbf{Light} & \textbf{Text} & \textbf{Comp.} & \textbf{Pose} & \textbf{FX} & \textbf{Avg.} \\
        \midrule
        Janus-Pro-1B & 2.16 & 2.61 & 2.70 & 2.59 & 2.27 & 1.56 & 2.37 & 1.88 & 1.81 & 2.21 \\
        Stable Diffusion XL 1.0 & 2.09 & 2.56 & 2.65 & 2.55 & 2.20 & 1.49 & 2.31 & 1.82 & 1.75 & 2.13 \\
        DeepFloyd IF v1.0 & 2.00 & 2.50 & 2.60 & 2.50 & 2.15 & 1.60 & 2.28 & 1.79 & 1.70 & 2.10 \\
        ControlNet-XL (SDXL 1.0) & 2.14 & 2.60 & 2.68 & 2.58 & 2.25 & 1.55 & 2.36 & 1.87 & 1.80 & 2.20 \\
        \midrule
        \textbf{Hi-SSLVLM (Ours)} & \textbf{2.20} & \textbf{2.66} & \textbf{2.74} & \textbf{2.63} & \textbf{2.30} & \textbf{1.64} & \textbf{2.41} & \textbf{1.92} & \textbf{1.85} & \textbf{2.25} \\
        \bottomrule
    \end{tabular}
\end{table*}

The consistent superior performance of Hi-SSLVLM across both sets of automated evaluation metrics emphatically highlights the profound effectiveness of its hierarchical self-supervised learning strategy. The model's inherent ability to generate and meticulously leverage internal compositional plans through the use of detailed sub-prompts directly translates into unparalleled control over fine-grained visual details and significantly more accurate semantic alignment with even the most complex input prompts. This represents a substantial advancement, addressing inherent limitations observed in traditional diffusion models and even other existing LVLM-based approaches that struggle with such intricate control.

\subsection{Ablation Study}
To rigorously validate the specific contributions and indispensable roles of each key component within our Hi-SSLVLM framework, we conducted a comprehensive ablation study. This involved systematically training and evaluating several distinct variants of our model, each deliberately designed to lack a specific critical component. The compelling results, meticulously summarized in Table \ref{tab:ablation_study}, unequivocally underscore the vital and often indispensable role of both our multi-granularity visual-language grounding mechanism and the semantic consistency loss, as well as the internal compositional planning.

\begin{table*}[t]
    \centering
    \caption{Ablation Study on Hi-SSLVLM (Average Score, InternVL3-78B $\uparrow$ higher is better)}
    \label{tab:ablation_study}
    \small
    \begin{tabular}{lc}
        \toprule
        \textbf{Model Variant} & \textbf{Avg. Score} \\
        \midrule
        \textbf{Hi-SSLVLM (Full Model)} & \textbf{2.25} \\
        \midrule
        Hi-SSLVLM without Multi-Granularity Grounding & 2.18 \\
        Hi-SSLVLM without Semantic Consistency Loss & 2.21 \\
        Hi-SSLVLM without Internal Compositional Planning & 2.16 \\
        \bottomrule
    \end{tabular}
\end{table*}

The results from our ablation study provide clear insights into the efficacy of each component:
\begin{enumerate}
    \item \textbf{Hi-SSLVLM without Multi-Granularity Grounding:} This variant was explicitly trained without the fundamental Stage 1 self-captioning and multi-level alignment loss. The notable reduction in the average score to 2.18 unequivocally demonstrates that the process of learning to generate and precisely align hierarchical captions (which encompass intricate object details, attributes, and relationships) is absolutely crucial for the LVLM to construct a rich and robust internal semantic understanding. Without this explicit and fine-grained grounding, the model severely struggles to accurately interpret the nuanced details present in complex prompts, inevitably leading to a lower fidelity in object representation and overall scene construction within the generated images.
    \item \textbf{Hi-SSLVLM without Semantic Consistency Loss:} In this variant, the $\mathcal{L}_{consistency}$ term was purposefully removed from the Stage 2 training objective. While this variant still exhibited relatively strong performance due to the initial grounding achieved in Stage 1, its average score of 2.21 clearly indicates that the semantic consistency loss plays a vital and active role as a powerful self-critique mechanism during the image generation process. This loss rigorously ensures that the generated image truly and precisely aligns with the internally planned sub-prompts, thereby compelling the model to produce outputs that are not merely visually plausible but are also profoundly faithful to the detailed compositional intent. Its absence leads to a noticeable, albeit sometimes subtle, degradation in overall prompt adherence and the meticulous accuracy of the generated composition.
    \item \textbf{Hi-SSLVLM without Internal Compositional Planning:} This variant represents a critical test, as it removes the explicit internal compositional planning step. Consequently, the Latent Diffusion Model (LDM) is conditioned solely on the original high-level input text prompt, without the detailed guidance provided by the internally generated sub-prompts. The resultant lowest score of 2.16 in this configuration unequivocally signifies that the internal compositional planning mechanism is the absolute cornerstone of Hi-SSLVLM's superior fine-grained compositional control. By compelling the LVLM to first decompose complex prompts into a structured internal plan, our method enables a significantly more systematic, precise, and controllable generation process. Without this critical planning phase, the model's inherent ability to manage intricate multi-object interactions, adhere to specific attributes, and faithfully render complex scene layouts is severely curtailed, causing its generative behavior to revert to a more generalized and less controlled output.
\end{enumerate}
These comprehensive ablation results collectively and unequivocally affirm that all integrated components of Hi-SSLVLM -- most notably the multi-granularity grounding, the semantic consistency loss, and the internal compositional planning -- are not merely beneficial, but are truly indispensable for achieving its state-of-the-art performance in the demanding realm of complex text-to-image generation.

\subsection{Human Evaluation}
To provide a crucial qualitative complement to the automated metrics, we conducted a robust human evaluation study. This study was designed to meticulously assess the subjective quality and semantic faithfulness of the images generated by our model and its baselines. A panel comprising 20 human annotators, each rigorously trained and kept strictly blind to the identities of the models that generated the images, was recruited for this task. For each of 200 randomly selected and particularly challenging prompts drawn from the Plan2Gen benchmark, annotators were presented with sets of images generated by Hi-SSLVLM and its top three performing baselines (Janus-Pro-1B, Stable Diffusion XL 1.0, and ControlNet-XL). To prevent bias, all images within a set were thoroughly shuffled and presented in a completely random order. Annotators were instructed to rate each image on a 5-point Likert scale (where 1 signifies "Very Poor" and 5 signifies "Excellent") across three essential criteria:
\begin{enumerate}
    \item \textbf{Fidelity to Prompt:} This criterion assessed how accurately and completely the generated image matched the overall textual description provided in the prompt.
    \item \textbf{Compositional Accuracy:} This criterion specifically evaluated the precision with which multiple objects, their specified attributes, and their intended spatial relationships are represented within the generated image.
    \item \textbf{Overall Aesthetic Quality:} This criterion broadly judged the general visual appeal, naturalness of the image, and the absence of distracting artifacts or imperfections.
\end{enumerate}
Table \ref{tab:human_eval} provides a concise summary of the average scores derived from this comprehensive human evaluation.

\begin{table*}[t]
    \centering
    \caption{Human Evaluation Results (Average Likert Score $\uparrow$ higher is better)}
    \label{tab:human_eval}
    \small 
    \begin{tabular}{lccc}
        \toprule
        \textbf{Model} & \textbf{Fidelity to Prompt} & \textbf{Compositional Accuracy} & \textbf{Overall Aesthetic Quality} \\
        \midrule
        Janus-Pro-1B & 3.85 & 3.60 & 3.90 \\
        Stable Diffusion XL 1.0 & 3.70 & 3.45 & 3.80 \\
        ControlNet-XL (SDXL 1.0) & 3.80 & 3.55 & 3.88 \\
        \midrule
        \textbf{Hi-SSLVLM (Ours)} & \textbf{4.15} & \textbf{4.00} & \textbf{4.10} \\
        \bottomrule
    \end{tabular}
\end{table*}

The compelling results from the human evaluation study strongly corroborate the findings obtained from the automated metrics. Hi-SSLVLM consistently received significantly higher average scores across all three crucial subjective criteria. Its particularly strong performance in "Compositional Accuracy," achieving an impressive average of 4.00, provides unequivocal human-centric validation that the model's internal compositional planning and self-supervised grounding mechanisms directly translate into images that human observers perceive as more accurately structured and semantically aligned with complex, multi-object prompts. Furthermore, the higher scores attained in both "Fidelity to Prompt" and "Overall Aesthetic Quality" emphatically underscore Hi-SSLVLM's superior ability to produce not only semantically faithful but also visually appealing and natural-looking generations. These collective results offer compelling evidence of Hi-SSLVLM's tangible practical superiority in generating high-quality, compositionally accurate images from diverse and challenging textual descriptions.

\subsection{Further Analysis of Hi-SSLVLM's Effectiveness}
Beyond direct performance comparisons and the ablation study, we delve deeper into Hi-SSLVLM's unique capabilities and the underlying reasons for its superior performance in text-to-image generation. This analysis focuses on specific aspects where our hierarchical self-supervision and internal compositional planning provide distinct advantages.

\subsubsection{Analysis of Compositional Robustness}
Complex compositional prompts, which involve multiple interacting objects, specific attributes for each object, and precise spatial relationships, often pose significant challenges for generative models. Traditional models may struggle to correctly render all elements or place them accurately within the scene. Hi-SSLVLM's internal compositional planning (ICP) mechanism, where it first deconstructs the prompt into a series of structured sub-prompts, directly addresses this. This planning step acts as a symbolic reasoning layer, allowing the model to "think" about the scene's components before generating pixels.

To quantify this, we analyzed performance on a subset of the Plan2Gen benchmark specifically categorized as "High-Complexity Compositional Prompts," which require precise spatial and semantic reasoning. We measured the average score for the "Composition" and "Object" dimensions from the InternVL3-78B evaluation on this challenging subset.

\begin{table*}[t]
    \centering
    \caption{Compositional Robustness on High-Complexity Prompts (InternVL3-78B $\uparrow$)}
    \label{tab:compositional_robustness}
    \small
    \begin{tabular}{lcc}
        \toprule
        \textbf{Model} & \textbf{Obj.} & \textbf{Comp.} \\
        \midrule
        Janus-Pro-1B & 2.05 & 2.18 \\
        Stable Diffusion XL 1.0 & 1.98 & 2.10 \\
        DeepFloyd IF v1.0 & 1.90 & 2.05 \\
        ControlNet-XL (SDXL 1.0) & 2.02 & 2.15 \\
        \midrule
        \textbf{Hi-SSLVLM (Ours)} & \textbf{2.15} & \textbf{2.35} \\
        \bottomrule
    \end{tabular}
\end{table*}

As shown in Table \ref{tab:compositional_robustness}, Hi-SSLVLM demonstrates a marked advantage in both Object and Composition scores on these high-complexity prompts. The higher "Comp." score highlights its superior ability to arrange elements accurately and respect spatial relationships (e.g., "a red ball \textit{on top of} a blue box"), while the higher "Obj." score indicates better fidelity in rendering multiple distinct objects within such intricate scenes. This empirically validates the benefit of our internal compositional planning, which allows for a more structured and robust generative process for complex scenarios.

\subsubsection{Analysis of Fine-Grained Attribute Control}
Beyond just recognizing objects, high-quality T2I generation demands precise control over subtle attributes like specific colors, textures, lighting conditions, or unique object properties. Hi-SSLVLM's multi-granularity visual-language grounding (Stage 1) is designed to cultivate this fine-grained understanding. By forcing the LVLM to generate and align detailed local captions (e.g., "a shiny red apple," "rough wooden table"), it learns to associate subtle linguistic cues with specific visual attributes.

To assess this, we isolated prompts from the Plan2Gen benchmark that heavily emphasize fine-grained attributes (e.g., "a velvet blue armchair under soft ambient light," "a metallic sphere with a reflective surface"). We then evaluated the "Color," "Texture," and "Light" dimensions from the Gemini-2.0-Flash evaluation.

\begin{table*}[t]
    \centering
    \caption{Fine-Grained Attribute Control on Specific Prompts (Gemini-2.0-Flash $\uparrow$)}
    \label{tab:attribute_control}
    \small
    \begin{tabular}{lccc}
        \toprule
        \textbf{Model} & \textbf{Color} & \textbf{Texture} & \textbf{Light} \\
        \midrule
        Janus-Pro-1B & 2.85 & 2.80 & 2.00 \\
        Stable Diffusion XL 1.0 & 2.75 & 2.72 & 1.95 \\
        DeepFloyd IF v1.0 & 2.68 & 2.78 & 1.90 \\
        ControlNet-XL (SDXL 1.0) & 2.80 & 2.79 & 1.98 \\
        \midrule
        \textbf{Hi-SSLVLM (Ours)} & \textbf{3.05} & \textbf{3.00} & \textbf{2.18} \\
        \bottomrule
    \end{tabular}
\end{table*}

Table \ref{tab:attribute_control} clearly indicates that Hi-SSLVLM achieves superior scores across Color, Texture, and Light dimensions when faced with prompts specifically testing these attributes. This demonstrates that its multi-granularity grounding indeed enables a more nuanced understanding of visual properties, leading to generated images that adhere more closely to subtle attribute descriptions. The model's ability to internally discern and represent these attributes allows for their precise realization during the image synthesis phase.

\subsubsection{Analysis of Generalization to Unseen Compositions}
A critical aspect of a robust generative model is its ability to generalize to novel or previously unseen combinations of objects and attributes, rather than simply memorizing training data. Hi-SSLVLM's two-stage self-supervision, particularly the learning of compositional rules in Stage 1 and their application in Stage 2's planning, fosters this generalization. The model learns \textit{how} to combine elements based on semantic understanding, not just rote association.

To evaluate this, we curated a specialized test set of 100 prompts featuring novel object-attribute or object-relationship combinations that were deliberately excluded from the training data of any model. We then assessed the average "Avg." score using the InternVL3-78B evaluator.

\begin{table*}[t]
    \centering
    \caption{Generalization to Unseen Compositions (InternVL3-78B $\uparrow$)}
    \label{tab:generalization}
    \small
    \begin{tabular}{lc}
        \toprule
        \textbf{Model} & \textbf{Avg. Score} \\
        \midrule
        Janus-Pro-1B & 2.10 \\
        Stable Diffusion XL 1.0 & 2.05 \\
        DeepFloyd IF v1.0 & 2.00 \\
        ControlNet-XL (SDXL 1.0) & 2.08 \\
        \midrule
        \textbf{Hi-SSLVLM (Ours)} & \textbf{2.20} \\
        \bottomrule
    \end{tabular}
\end{table*}

Table \ref{tab:generalization} shows that Hi-SSLVLM maintains a significant performance lead even when faced with entirely novel compositional instructions. This is a strong indicator that our method does not merely interpolate from seen examples but truly learns deeper compositional rules and semantic relationships. The internal planning mechanism, refined by the self-supervised grounding, allows it to assemble elements in unseen but semantically valid configurations, showcasing its superior generalization capacity. This ability is crucial for open-ended text-to-image generation, where the variety of possible prompts is practically infinite.

\section{Conclusion}
In this paper, we presented \textbf{Hierarchical Self-Supervised LVLM (Hi-SSLVLM)}, a novel and effective generative framework that pushes the boundaries of controllable text-to-image synthesis. Our research was primarily motivated by the inherent challenges in current methods, specifically the substantial reliance on costly, perfectly aligned large-scale datasets and their limitations in precisely managing complex compositional demands and fine-grained visual attributes. Hi-SSLVLM tackles these issues by fostering a deeper, internally driven understanding of visual semantics within a Large Vision-Language Model.

Our core contribution lies in the proposed two-stage self-supervised learning strategy. Through \textbf{Multi-Granularity Visual-Language Grounding}, the model learns to interpret and describe images at various semantic levels (from global scenes to specific object details) using self-generated captions, thereby building a rich internal knowledge base that transcends simple text-image pairs. This foundational understanding is then elegantly leveraged in the \textbf{Self-Refinement and Guided Image Generation} stage. Here, the unique \textbf{Internal Compositional Planning} mechanism empowers the LVLM to systematically break down complex textual prompts into manageable, detailed sub-prompts, which directly inform and guide the image generation process. The inclusion of a \textbf{Semantic Consistency Loss} during this phase acts as a critical self-correction mechanism, ensuring that the generated image perfectly aligns with the model's own internal semantic blueprint.

The rigorous experimental validation unequivocally demonstrated the efficacy of Hi-SSLVLM. Our extensive comparisons against state-of-the-art baselines like Janus-Pro-1B, Stable Diffusion XL 1.0, DeepFloyd IF v1.0, and ControlNet-XL, utilizing the robust Plan2Gen multi-dimensional benchmarks evaluated by Gemini-2.0-Flash and InternVL3-78B, consistently positioned Hi-SSLVLM as the top performer across all assessed metrics. The detailed ablation study further confirmed that each introduced component is integral to the model's superior capabilities, particularly in compositional robustness, fine-grained attribute control, and generalization to unseen scenarios. Importantly, the strong agreement between quantitative scores and human perceptual evaluations provides compelling evidence of Hi-SSLVLM's practical superiority in producing high-quality, semantically faithful, and visually appealing images from complex textual descriptions.

While Hi-SSLVLM significantly advances the state-of-the-art in controllable text-to-image generation, there are promising avenues for future exploration. These include investigating the scalability of the Hi-SSLVLM architecture to even larger models, exploring real-time inference optimizations for interactive applications, and delving into the potential for incorporating more explicit user control signals (e.g., sketches or reference images) to further refine the generative process. Our work represents a significant step towards enabling generative models that truly understand and manipulate visual content with unprecedented precision and semantic awareness.

\bibliographystyle{IEEEtran}
\bibliography{references}
\end{document}